%% file: aaai25.tex
\documentclass[letterpaper]{article} % DO NOT CHANGE THIS
\usepackage{aaai25}  % DO NOT CHANGE THIS
\usepackage{times}  % DO NOT CHANGE THIS
\usepackage{helvet}  % DO NOT CHANGE THIS
\usepackage{courier}  % DO NOT CHANGE THIS
\usepackage[hyphens]{url}  % DO NOT CHANGE THIS
\usepackage{graphicx} % DO NOT CHANGE THIS
\usepackage{natbib}  % DO NOT CHANGE THIS AND DO NOT ADD ANY OPTIONS TO IT
\usepackage{caption} % DO NOT CHANGE THIS AND DO NOT ADD ANY OPTIONS TO IT
\usepackage{algorithm}
\usepackage{algorithmic}
\usepackage{subcaption}
\usepackage{tabularx}
\usepackage{array}
\usepackage{multirow}
\usepackage{longtable}
\usepackage{array}
\usepackage{booktabs}
\usepackage[toc,page]{appendix}
\usepackage{todonotes}
\usepackage[section]{placeins}

\usepackage{newfloat}
\usepackage{listings}
\DeclareCaptionStyle{ruled}{labelfont=normalfont,labelsep=colon,strut=off} % DO NOT CHANGE THIS
\floatstyle{ruled}
\newfloat{listing}{tb}{lst}{}
\floatname{listing}{Listing}
\title{Towards Unifying Evaluation of Counterfactual Explanations: Leveraging Large Language Models for Human-Centric Assessments\thanks{The Thirty-Ninth AAAI Conference on Artificial Intelligence (AAAI-25)}}
\author {
    Marharyta Domnich\textsuperscript{\rm 1},
    Julius Välja\textsuperscript{\rm 1},
    Rasmus Moorits Veski\textsuperscript{\rm 1,2},
    Giacomo Magnifico\textsuperscript{\rm 1},
    Kadi Tulver\textsuperscript{\rm 1},
    Eduard Barbu\textsuperscript{\rm 1},
    Raul Vicente\textsuperscript{\rm 1}%corresponding author
}
\affiliations{
    \textsuperscript{\rm 1} Institute of Computer Science, University of Tartu, Tartu, Estonia\\
    \textsuperscript{\rm 2} École Polytechnique Fédérale de Lausanne, Lausanne, Switzerland
    \textrm{marharyta.domnich@ut.ee}
}

\begin{document}

\maketitle

\begin{abstract}
As machine learning models evolve, maintaining transparency demands more human-centric explainable AI techniques. Counterfactual explanations, with roots in human reasoning, identify the minimal input changes needed to obtain a given output, and hence, are crucial for supporting decision-making. Despite their importance, the evaluation of these explanations often lacks grounding in user studies and remains fragmented, with existing metrics not fully capturing human perspectives. To address this challenge, we developed a diverse set of 30 counterfactual scenarios and collected ratings across 8 evaluation metrics from 206 respondents. Subsequently, we fine-tuned different Large Language Models (LLMs) to predict average or individual human judgment across these metrics. Our methodology allowed LLMs to achieve an accuracy of up to 63\% in zero-shot evaluations and 85\% (over a 3-classes prediction) with fine-tuning across all metrics. The fine-tuned models predicting human ratings offer better comparability and scalability in evaluating different counterfactual explanation frameworks. 
\end{abstract}

 %\begin{links}
     %\link{Code}{https://github.com/anitera/CounterEval}
    %\link{Dataset}{https://huggingface.co/datasets/anitera/CounterEval}
     %\link{Appendix}{https://arxiv.org/abs/2410.21131}
 %\end{links}

\section{Introduction}
%\begin{itemize}
%    \item Counterfactual explanations are important
%    \item Every method use their own metrics
%    \item Those metrics overlook human aspect and there are multiple explanatory virtues
%    \item Human studies are biased and necessity for benchmark
%    \item LLMs potential
%\end{itemize} 

The rapid adoption of AI in various domains has significantly increased the urgency for explainable AI models. Counterfactual explanations, which address the question "How should the input be different in order to change the model's decision outcome?" \cite{wachter2017counterfactual}, not only clarify the machine's reasoning but also suggest potential changes that users might implement to achieve different results. These explanations enhance user trust and understanding by providing a richer mental representation compared to causal explanations \cite{warren2023categorical}. Additionally, counterfactual explanations align closely with human cognitive processes \cite{miller_explanation_2019}, as they provide alternative hypothetical realities that are pervasive in our natural reasoning \cite{byrne2002mental}.

Evaluating counterfactual explanations poses a significant challenge in the field. While various quantitative metrics, such as validity, proximity, sparsity, coherence, robustness, and diversity \cite{guidotti_counterfactual_2022, karimi2022survey,rasouli2024care} are currently used, they often fall short in capturing the human perspective, missing key explanatory virtues, and leading to inconsistent findings that complicate the development of a standardized evaluation framework. User studies are commonly recommended to assess the efficacy of counterfactual explanations, as "excellent computational explanations may not be good psychological explanations" \cite{keane2021bettercounterfactualexplanationskey}. Despite this, such studies are rarely utilized for benchmarking counterfactual explanations \cite{LONGO2024102301}. One of the reasons for this is the difficulty and expense of recruiting a sufficient number of experts capable of performing these evaluations. Even when executed, user studies do not guarantee consistent and reproducible results as perceptions of what constitutes a reasonable explanation can vary widely between individuals and user groups \cite{KENNY2021103459}. Furthermore, most studies only employ a few qualitative measures, such as satisfaction and trust, which fail to address the nuanced features influencing human preferences \cite{warren2023categorical}. While human assessments of counterfactual explanations are invaluable, these issues of cost and scalability make it very challenging to make meaningful comparisons and generalizations between multiple frameworks or domains.  

\begin{figure*} [ht]
\begin{center}
\includegraphics[width=\textwidth]{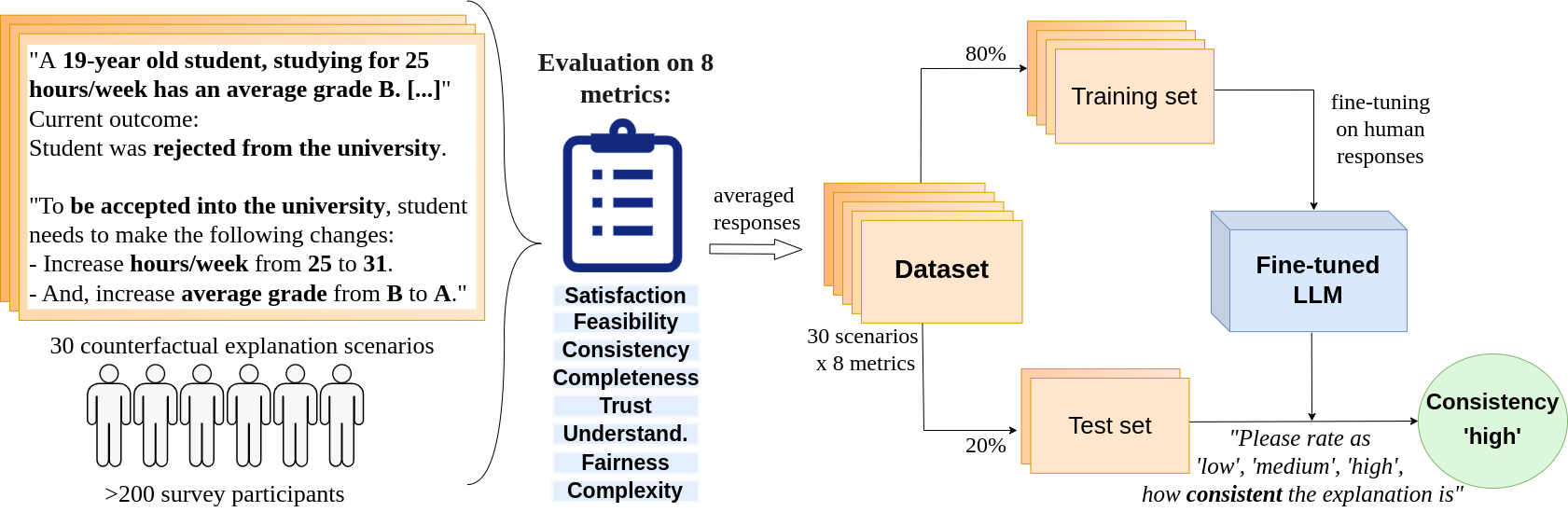}
\caption{We created a diverse set of counterfactual scenarios where we varied feasibility, consistency, completeness, trust, fairness, complexity and understandability, resulting in 30 counterfactual questions which were evaluated by 206 human respondents on the 8 metrics. We subsequently divided data for fine-tuning several LLM models to assess every metric score and compared the results to human data on a reserved set.}
\label{fig:pipeline}
\end{center}
\end{figure*}

Recognizing the limitations of existing methodologies, this paper explores the potential of Large Language Models (LLMs) to serve as a benchmark for automating the evaluation of counterfactual explanations. Current LLMs have demonstrated remarkable capabilities in interacting with natural language data, ranging from extensive data summarisation \cite{liu2024trustllm} and pattern deduction \cite{jin2024-llmnlptimeseries} to idea generation \cite{llm-ideagen} and problem-solving through branching solutions \cite{yang2024-llmoptim}. Based on these premises, LLMs are hypothesized to mimic human evaluative judgments effectively, offering a more accessible and cost-efficient alternative to traditional methods.

%LLMs, which have demonstrated remarkable capabilities in understanding and generating text \cite{Wang2024-survey}, are hypothesized to mimic human evaluative judgments effectively, offering a more accessible and cost-efficient alternative to traditional methods. 

In light of these considerations, this paper addresses the following question: \textbf{Can the evaluation process of counterfactual explanations be effectively automated using LLMs?} To answer this question, we created a diverse set of 30 counterfactual scenarios that were designed to vary across multiple dimensions of explanatory qualities. The scenarios were evaluated by 206 human respondents in overall satisfaction, feasibility, consistency, completeness, trust, fairness, complexity, and understandability. Next, we divided data for fine-tuning several LLM models to assess every metric score and compared the results to human data on a reserved test set. The pipeline can be seen in Figure~\ref{fig:pipeline}.

%By integrating insights from extensive literature and empirical data, this paper addresses the following critical questions:

%Can the process of evaluating counterfactual explanations be automated using LLMs?
%How do LLM evaluations compare to human judgments in terms of accuracy and satisfaction?
%What are the key metrics that effectively predict human satisfaction with counterfactual explanations?
Through systematic exploration, this study seeks to bridge the gap between algorithmic outputs and human-centric evaluations, advancing towards more reliable and universally accepted counterfactual explanations in AI systems.

The contributions of the paper are twofold:
\begin{itemize}
    \item First, we present a diverse dataset of human-evaluated counterfactual explanations CounterEval, encompassing a variety of metrics and scenarios, which could serve both for benchmarking and for training better causal representations of data, as demonstrated in \cite{chen2023discodistillingcounterfactualslarge}.
    \item Second, we introduce a fine-tuned LLM-based evaluator of counterfactual explanations that captures understanding of various explanatory virtues, such as Feasibility, Consistency, Trust, Completeness, Understandability, Fairness, Complexity and Overall Satisfaction.
\end{itemize}

\section{Related Works}
In the following section, we review user studies that focus on evaluating counterfactual explanations, and the potential of LLMs to simulate human responses. 
\subsection{User Studies for Evaluating Counterfactual Explanations}
In addition to quantitative explanatory metrics like proximity, validity, or sparsity, most researchers agree that it is crucial to capture the subjective preferences of human users to achieve more human-centric AI explanations \cite{Kirsch2017ExplainTW,keane2021bettercounterfactualexplanationskey,LONGO2024102301}. Yet, a survey found that only 21\% of 100 studies on counterfactual methods included user evaluations  \cite{keane2021bettercounterfactualexplanationskey}. Furthermore, many of those studies test the use of counterfactual explanations vs no-explanations rather than comparing different methods, leaving only 7\% of papers that report user evaluations for benchmarking different counterfactual algorithms. 

In recent years, some user studies have been conducted with tabular counterfactual data. For instance, \cite{warren2023categorical} conducted a study with 127 participants to compare the effects of counterfactual and causal explanations on objective prediction accuracy and subjective judgments of satisfaction and trust. \cite{bove2023investigating} explored the impact of plural counterfactual examples on objective understanding and a modified Explanation Satisfaction Scale \cite{hoffman2018metrics} in a lab study with 112 participants. \cite{forster2021capturing} conducted a study with 46 participants assessing explanation realism and typicality. Two user studies have benchmarked counterfactual methods for perceived practicality of users in a study with 135 participants \cite{ghazimatin2020prince}, and an online study with 500 responders \cite{spreitzer2022evaluating}. Additionally, \cite{AKULA2022103581} tested their approach on image data, evaluating justified trust as a quantitative metric and explanation satisfaction as qualitative metric. 

Overall, user studies on explanation satisfaction often focus on a limited range of aspects \cite{mueller2019explanation}, typically measuring satisfaction and trust, while neglecting other essential qualities of the explanations themselves. Such studies may fail to capture human preferences, which are shaped by context, presentation, and cognitive biases, especially when preferences are ill-defined \cite{kliegr2021review, tversky1993context}. As a result, this limited scope leads to inconsistent perspectives on explanatory qualities, leaving a significant gap in understanding which features define good explanations. 

\subsection{Potential of LLMs in Simulating Human Responses}

Predicting human evaluation with machine learning has gained widespread acceptance in various domains, such as human-computer interaction \cite{10.1145/2911451.2911521}, recommendation systems \cite{10.1145/3624989}, speech quality assessment \cite{reddy2022dnsmos}, etc. 
The advancement of LLMs' causal reasoning abilities \cite{llm-counterfactual-text} supports their use in explainability, as their natural language explanations exhibit qualities similar to human output \cite{augmenting_xai_llms_castelnovo_2024} and the explanatory process can be further enhanced through a post-output chat pipeline \cite{Slack2023}.
%Although there are currently available strategies to deploy multiple large models as teams of autonomous agents in order to generate answers with expert-specific knowledge, even single LLMs provide a precedent for data expertise being retained and presented to the user without requiring fine-tuning \cite{wang-etal-2024}.
LLMs have also been used to evaluate and model user satisfaction, providing insight into choices and preferences \cite{kim2024usingllmsinvestigatecorrelations}, and as artificial user-model tuning pairs \cite{gao2024-agentusermodel}. 

At the time of this paper's submission, no prior work existed on simulating human assessments for evaluating counterfactual explanations using LLMs. However, a concurrent effort has since emerged, replicating an existing user study that compared counterfactual and causal explanations by replacing human participants with seven LLMs \cite{debona2024evaluatingexplanationsllmstraditional},  demonstrating that LLMs can replicate some conclusions from the original study. Nonetheless, to the best of our knowledge, no work has yet focused on evaluating the intrinsic quality of counterfactual explanations.

\section{Development and Human Evaluation of a Counterfactual Explanation Dataset}
Training LLMs to evaluate the quality of counterfactual explanations in a human-like manner requires human-labeled data. Currently, no widely-used dataset of human-evaluated counterfactual explanations exists. To fill this gap, we created a varied dataset of 30 counterfactual explanation instances, which were graded on 8 different criteria by 206 people through an online survey.

\subsection{Dimensions of Explanatory Qualities}

To select the dimensions for our study, we reviewed the literature on qualitative metrics that influence human judgments. Among the most frequently cited explanatory virtues are coherence and simplicity \cite{mackonis2013inference}, aligning with the understanding of human mental models and a preference for consistent and parsimonious information \cite{johnson2010mental}. 

\textbf{Coherence} can be measured internally, representing consistency within the explanation, or externally, taking into account the prior knowledge of the rater \cite{zemla2017evaluating}. Our work focuses on internal coherence, measuring consistency within different parts of the explanation, independent of an individual's prior experiences. 

The virtue of simplicity, also referred to as (Desired) \textbf{Complexity} \cite{zemla2017evaluating} or Selection \cite{VILONE202189}, assumes that people prefer simple explanations \cite{lombrozo2007simplicity}. However, evidence suggests humans sometimes favor complex explanations involving more causal links, or that moderate complexity and sufficient detail are preferred \cite{zemla2017evaluating, hoffman2018metrics}. In this study, we include Complexity as a metric, with desired values lying in the middle, as explanations may be perceived as overly simple or complex.

A commonly assessed quality in user studies is \textbf{Trust}. Definitions often focus on trust in the system generating explanations \cite{perrig2023trust}. Trust in explanations is considered in terms of trustworthiness, or the perceived credibility of suggested changes \cite{STEPIN2022379}. We define Trust as the belief that following the explanation will lead to the desired outcome.

\textbf{Feasibility} is one of the most agreed-upon metrics when discussing counterfactual explanations, although discussed under different names: Controllability~\cite{Byrne2019}, Actionability~\cite{rasouli2024care} and split into Actionability and Mutability~\cite{karimi2022survey}. While actionability has been employed as a quantitative measure \cite{guidotti_counterfactual_2022}, feasibility refers to whether the proposed changes are perceived as achievable and realistic. Explanations that fail this criterion are rated poorly \cite{butz2024evaluating}. 

\textbf{Understandability}, also known as Readability \cite{STEPIN2022379} or Comprehensibility \cite{VILONE202189}, relates to how effectively an explanation conveys the model’s decision process to the user and how easily it is grasped. Higher understandability is generally linked to greater user satisfaction, with clear explanations preferred, though complex answers may be favored in some contexts. 

\textbf{Completeness} previously discussed as Incompleteness \cite{zemla2017evaluating} or Informativeness, the latter of which also includes the notion of extraneous information \cite{STEPIN2022379}, is tied to understanding causal relations and partially depends on domain knowledge \cite{keil2006explanation}. Evaluating completeness is challenging, since people often fill logical gaps in explanations \cite{strickland2011event}.

Finally, the dimension of \textbf{Fairness} in counterfactual explanations has been highlighted in recent work \cite{wang2024counterfactual}. Concerns about models that unintentionally encode or amplify biases in training data \cite{corbett2023measure} make it crucial to address potential unfairness and discrimination. Fairness has been viewed mainly as a quantitative metric \cite{ge2022explainable}, with limited understanding of its influence on the perceived quality of the explanation.

\subsection{Generating Counterfactual Explanations Scenarios}

Relying on previous work on human preferences and explanatory virtues, we selected 8 different criteria capturing a range of relevant dimensions (see the previous section for an overview) to guide the creation of diverse counterfactual scenarios. The scenarios in our study are grounded in actual outputs from counterfactual algorithms applied to commonly used datasets for counterfactual explanation evaluation, such as the Adult dataset and the Pima Indians Diabetes dataset. In some cases, we modified the algorithmic outputs to ensure a broader representation across explanatory quality metrics. Additionally, we tailored certain features to enhance clarity for human evaluators based on feedback from a pilot study. All counterfactual scenarios were designed from the perspective of improving the factual situation, as directionality has been shown to influence the way explanations are perceived \cite{kuhl2023better}.

We included examples of explanations that fulfilled the different qualities at varying levels to train LLM models to distinguish between good and bad explanations. Specific instances were created by varying metrics, with the exception of \textbf{Understandability} and \textbf{Overall satisfaction}. We did not specifically vary the overall satisfaction of explanations, as this metric serves as a general indicator of the perceived quality of an explanation. Also, all explanations were designed to be as understandable as possible, and no instances with purposefully poor wording were included to ensure participants could reliably assess other metrics.

Our dataset contained examples of extreme changes in both categorical and continuous features, as people may evaluate these differently \cite{warren2023categorical}. For example, we explored how humans perceive \textbf{Feasibility} by creating explanations which changed inactionable features (e.g. age); features by different margins (a 1000€ pay increase vs 10 000€); continuous features outside and within distribution, starting from the value 0; and ordinal features in infeasible directions (e.g. lowering education level). For \textbf{Consistency}, we changed features widely considered connected (e.g. hours studied and average grade) in both covarying and conflicting directions, using categorical and continuous features. Differences in \textbf{Completeness} were implemented with sufficiently detailed explanations or those containing obvious gaps. Furthermore, useful context was provided for some questions to ensure minimal domain knowledge was sufficient, the lack of which could influence perceptions of completeness. Variety in \textbf{Trust} was induced by having logical, solution-oriented explanations and others unlikely to bring about the desired change. Poor \textbf{Fairness} was represented by recommendations involving controversial features  (e.g. gender, age). To vary \textbf{Complexity}, we included instances that might be perceived as too complex as well as too simple by having explanations with a different length and number of recommendations to similar problems. Here, we hypothesised that a desired level of Complexity lies in the middle, which is also reflected in the slightly different scale of measurement compared to other metrics. All selected metrics, along with their definitions and scales as presented in the questionnaire, are detailed in Table~\ref{tab:metric_definitions}.

\begin{table}[ht]
\begin{tabular}{|p{0.25\columnwidth}|p{0.6\columnwidth}|}
\hline
\textbf{Metric and scale}                                            & \textbf{Description}                                                                                        \\ \hline
\begin{tabular}[t]{@{}p{2cm}@{}}\textbf{\textit{Overall}}\\ \textbf{\textit{satisfaction}}\\ from 1 to 6\end{tabular} & This scenario effectively explains how to reach a different outcome                                        \\ \hline
\begin{tabular}[t]{@{}p{2cm}@{}}\textbf{\textit{Feasibility}}\\ from 1 to 6\end{tabular}          & The actions suggested by the explanation are practical, realistic to implement and actionable              \\ \hline
\begin{tabular}[t]{@{}p{2cm}@{}}\textbf{\textit{Consistency}}\\ from 1 to 6\end{tabular}          & All parts of the explanation are logically coherent and do not contradict each other                       \\ \hline
\begin{tabular}[t]{@{}p{2cm}@{}}\textbf{\textit{Completeness}}\\ from 1 to 6\end{tabular}         & The explanation is sufficient in explaining the outcome                                                    \\ \hline
\begin{tabular}[t]{@{}p{2cm}@{}}\textbf{\textit{Trust}}\\ from 1 to 6\end{tabular}                & I believe that the suggested changes would bring about the desired outcome                                 \\ \hline
\begin{tabular}[t]{@{}p{2cm}@{}} \textbf{\textit{Understand.}}\\ from 1 to 6\end{tabular} & I feel like I understood the phrasing of the explanation well \\ \hline
\begin{tabular}[t]{@{}p{2cm}@{}}\textbf{\textit{Fairness}}\\ from 1 to 6\end{tabular}             & The explanation is unbiased towards different user groups and does not operate on sensitive features       \\ \hline
\begin{tabular}[t]{@{}p{2cm}@{}}\textbf{\textit{Complexity}}\\ from -2 to 2\end{tabular}          & The explanation has an appropriate level of detail and complexity - not too simple, yet not overly complex \\ \hline
\end{tabular}
\caption{Definitions of the evaluation criteria provided to the respondents in the questionnaire with ranking scale (Understand. stands for Understandability).}
\label{tab:metric_definitions}
\end{table}

\subsection{Questionnaire Results}
To assess the suitability and comprehensibility of the compiled scenarios and evaluation metrics, a pilot study was conducted with 15 volunteers recruited from university students and colleagues. Feedback gathered during the pilot led to revisions in the wording of some metric descriptions. Additionally, the Coherency metric was renamed to Consistency and Bias was changed to Fairness to aid comprehension for the participants.

The final version containing 30 counterfactual scenarios was shared on the Prolific platform and evaluated by 206 respondents. On average, completing the questionnaire took 42 minutes. All metrics were rated on scales detailed in Table~\ref{tab:metric_definitions}, with a 6-point ordinal scale from 1 (lowest) to 6 (highest) except for Complexity, rated on a 5-point scale from -2 (too simple) to 2 (too complex), where 0 corresponded to desired complexity. Counterfactual scenarios were presented to participants in random order, while the evaluation metrics remained in the same order. All respondents had to be at least 18 years of age and fluent in English to participate. CounterEval dataset with annotated human responses is available on HuggingFace\footnote{https://huggingface.co/datasets/anitera/CounterEval}, and an example question is shown in Appendix A, Table A.1 (refer to the extended version of this paper for appendix \cite{domnich2024unifyingevaluationcounterfactualexplanations}).

To detect fraudulent participants, a hidden attention check question was included in the questionnaire. Responses were also analyzed based on response time, average understandability score, respondence clustering, and the uniformity of response patterns. Additionally, individual answers to 3 indicator questions were reviewed. For example, if a participant rated an explanation recommending a change in place of birth as feasible, that respondent was flagged. Respondents failing the three aforementioned criteria were excluded from further analysis, with a total of 10 respondents removed.

The survey results indicated satisfactory variance in ratings of the metrics. The questionnaire contained examples of extreme ratings for all metrics with the mean usually balanced in the middle of the scale, as seen in Table~\ref{tab:descriptive_statistics}.
\begin{table}[ht]
\centering
\begin{tabular}{|c|c|c|}
\hline
\textbf{Metric}& \textbf{mean (\textpm sdv)} & \textbf{min / max}  \\
             \hline
Satisfaction & 3.02 (\textpm 1.11) & 1.4 / 5.21 \\
\hline
Feasibility  & 3.27 (\textpm 1.15) & 1.34 / 5.11 \\
\hline
Consistency  & 3.69 (\textpm 1.14) & 1.77 / 5.43 \\
\hline
Completen.   & 3.38 (\textpm 0.92) & 1.78 / 5.33 \\
\hline
Trust        & 3.16 (\textpm 1.15) & 1.42 / 5.32 \\
\hline
Understand.  & 4.82 (\textpm 0.51) & 3.92 / 5.58 \\
\hline
Fairness     & 3.89 (\textpm 0.97) & 1.61 / 5.42 \\
\hline
Complexity   & -0.26 (\textpm 0.39) & -1.03 / 0.84 \\ 
\hline
\end{tabular}
\caption{Metric statistics with values averaged per individual question. The table displays mean, standard deviation (sdv), minimum (min), and maximum (max) values.}
\label{tab:descriptive_statistics}
\end{table}

\begin{figure} [ht]
\begin{center}
\includegraphics[width=\columnwidth]{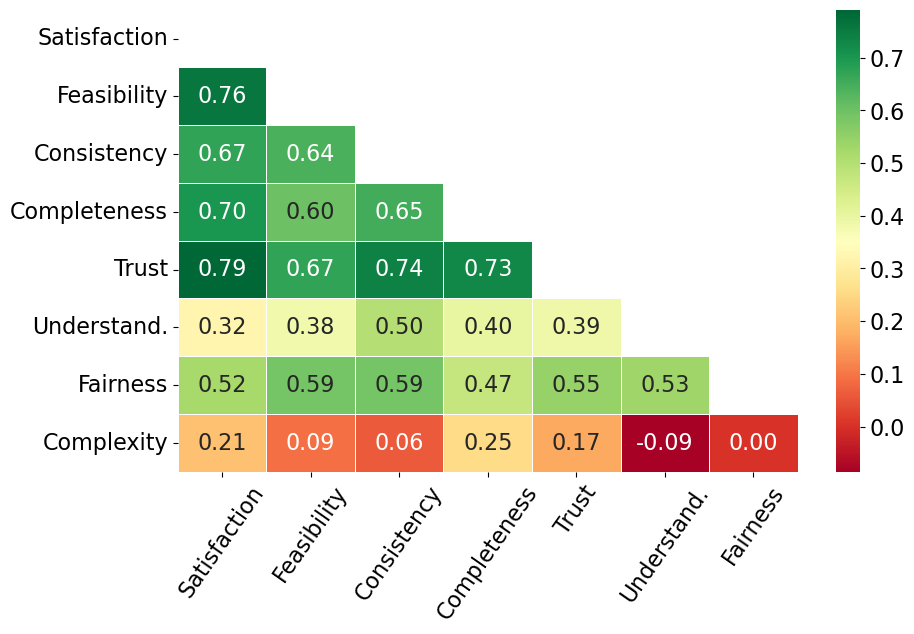}
\caption{Spearman correlation table between metrics. The values for Complexity were mapped linearly from the original [-2,2] scale to [1,6] to be in line with the other metrics.}
\label{fig:spearman_corr}
\end{center}
\end{figure}

The correlation diagram in Figure~\ref{fig:spearman_corr} shows that all explanatory qualities significantly correlate with each other (p-value $< 10^{-4}$, $\alpha$ = $\frac{0.05}{28}$), except for Complexity and Fairness. An analysis of questions involving varied fairness revealed they did not include overly complex explanations. The intercorrelated responses are likely to reflect that humans grade the explanations as a whole, rating different metrics in the context of the entire scenario and other explanatory virtues. Notably, all metrics correlate positively with satisfaction, highlighting their importance for evaluating the overall quality of counterfactual explanations. Furthermore, reducing the 7 metrics' scores (excluding Overall Satisfaction) to a two-dimensional space using t-SNE, and coloring by Satisfaction, shows a distinct distribution correlating with overall satisfaction, as detailed in Appendix A, Figure A.1.

\section{Modelling Human Assessment With LLMs}
With the questionnaire data as the input dataset, we aimed to test and fine-tune LLMs for automated evaluation of counterfactual explanations. The models selected for this were Llama 3.1 Instruct, Llama 3 Instruct \cite{dubey2024llama3herdmodels} and GPT-4 \cite{openai2024gpt4}. Llama models were fine-tuned on HPC clusters equipped with NVIDIA Tesla A100 GPUs using the transformers library by Huggingface (Wolf et al.
2020). QLoRA, which relies on rank decomposition matrices and quantization, was used to reduce memory requirements during fine-tuning ~\cite{dettmers_qlora_2023}.

\subsection{Dataset Preparation}
After gathering and filtering questionnaire responses, further data processing was needed. The preprocessing scripts and model weights are publicly available in the GitHub repository\footnote{https://github.com/anitera/CounterEval}. For each question-metric pair, the average response from 196 participants was used as the final value.  Complexity, originally rated on a -2 to 2 scale, was linearly scaled to align with the 1 to 6 scale used for other metrics. To minimize scale effects and enhance generalizability, we consolidated all metric values into three distinct categories: "low," "medium," and "high". Data analysis suggested that the differences between scores of 1 and 2, 3 and 4, and 5 and 6 could be effectively compressed into these categories. This classification ensured a balanced distribution across the dataset. With 30 questions and 8 metrics per question, this resulted in 240 instances of metric evaluation in total.

\subsection{Prompt Engineering}
To achieve the best possible performance from an LLM, three prompt structures were tested and compared. 

Importantly, the instruction part of the prompt was taken from the questionnaire directly to ensure that the task reflects the gathered data, and all changes were made in what is known as a “system prompt”. The following system prompts were developed:

\begin{itemize}
    \item A baseline prompt which contains an introduction to counterfactual explanations, the expected output format, and the definition of the metric being evaluated.
    \item A prompt that contains all the information present in the baseline prompt, but additionally provides definitions for all the metrics, not just the metric being evaluated.
    \item A prompt that additionally contains two examples of input and expected output, one with Consistency rated as “high” and the other with Feasibility rated as “low”. These examples were crafted based on the examples provided for metrics in the questionnaire. The specific examples were chosen to contain different metrics and different output values. All the additional information present in previous prompts is contained in this prompt as well.
\end{itemize}

The instruction or “user prompt” was adapted from the questionnaire, meaning it contained a factual-counterfactual pair from the questionnaire, alongside a modified metric evaluation question, such as "Please rate as 'low' (very unfeasible), 'medium' or 'high' (completely feasible), how feasible is this explanation:”. Consequently, each counterfactual explanation resulted in 8 instances, one for every metric under evaluation. The specific phrasing of all prompts can be found in Appendix B.

All of the prompts were tested using preliminary data from 100 participants and four LLMs, including Mistral-7B Instruct, Llama 2 7B Chat, and 8B and 70B versions of Llama 3 Instruct. Based on the results ( Appendix B, Table B.1), the baseline prompt was selected for all further experiments.

\subsection{Modelling Averaged Human Ratings}
Two data splits were tested, with 20\% of the dataset set aside for testing and 80\% used for training LLMs. The first experiment used a metric-based split, ensuring the testing dataset contained examples from all metrics in equal amounts, with 6 examples per metric. In addition, it provided at least one example with a 'high', 'medium' and 'low' answer for every metric. This split has the advantage of a bigger set of unique counterfactual explanation scenarios being present in the test set, leading to a more diverse range of metrics.

The second split, focused on counterfactual explanations, comprised 6 hand-picked questions for the test set. Each question was initially designed to assess a specific metric, typically aiming to elicit either a positive or negative evaluation of that metric. This design informed the selection of questions for the testing set, ensuring that each question covered a different metric with both positive and negative examples. This split accounts for correlations between metrics and ensures that none of the questions are shown in the training set associated with different metrics. 

\begin{table}[ht]
\centering
% Define the width of each column manually
\begin{tabular}{|>{\centering\arraybackslash}p{2cm}|>{\centering\arraybackslash}p{1cm}|>{\centering\arraybackslash}p{1cm}|>{\centering\arraybackslash}p{1cm}|>{\centering\arraybackslash}p{1cm}|}
\hline
\multirow{3}{*}{\textbf{Model}} & \multicolumn{2}{c|}{\textbf{Metric Split}} & \multicolumn{2}{c|}{\textbf{Question Split}} \\ \cline{2-5} 
 & Zero- & Fine- & Zero- & Fine- \\
 & shot & tuned & shot & tuned \\
\hline
Llama 3 8B & 0.48 & 0.80 & 0.45 & 0.77 \\ 
\hline
Llama 3.1 8B & 0.52 & \textbf{0.85} & 0.50 & 0.74 \\ 
\hline
GPT-4 & \textbf{0.63} & - & 0.58 & - \\ 
\hline
Llama 3 70B & 0.57 & \textbf{0.85} & \textbf{0.59} & \textbf{0.81} \\ 
\hline
\end{tabular}
\caption{Accuracy for metric-based and question-based testing set across evaluated LLMs.
Scores averaged over 4 runs, highest score for each column highlighted in bold.}
\label{tab:combined_results}
\end{table}

The optimal fine-tuning hyperparameters for every model were discerned through extensive testing (see Appendix C, Table C.1). All models were fine-tuned using a completion-only data collator from Huggingface’s \textit{trl} library~\cite{werra_trl_2020} to improve the predictive performance of the models. With a typical language modeling data collator, the model would have learned to predict the question text as well, but this was unnecessary for the task at hand. Due to its proprietary nature, GPT-4 was not used for fine-tuning.

\begin{figure}[ht]
    \centering
    % First row of subfigures
    \begin{subfigure}[b]{0.48\linewidth}
        \includegraphics[width=\linewidth]{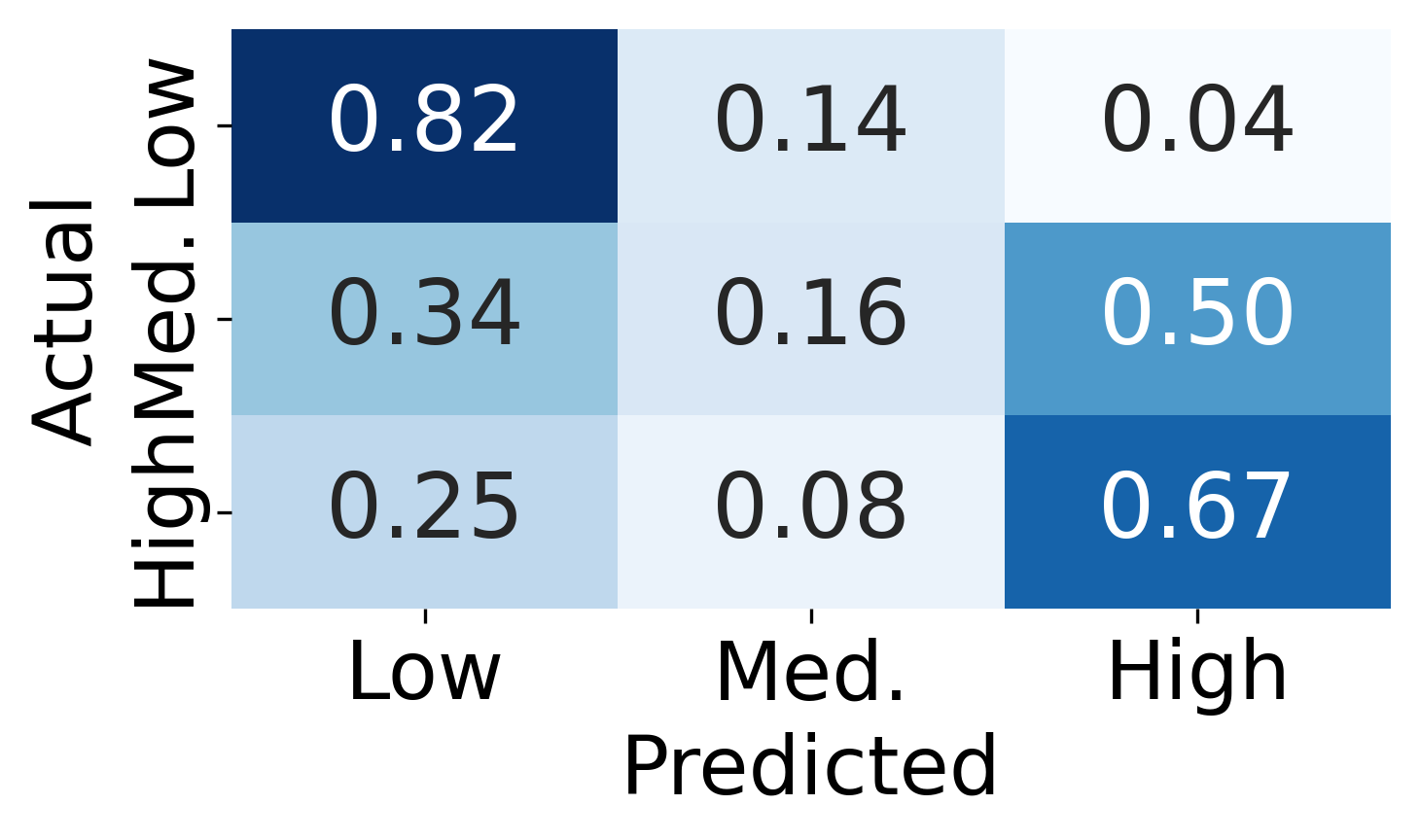}
        \caption{zero-shot metric split}
        \label{fig:cf_zh_metric}
    \end{subfigure}
    \hfill % Spacing between the first row subfigures
    \begin{subfigure}[b]{0.48\linewidth}
        \includegraphics[width=\linewidth]{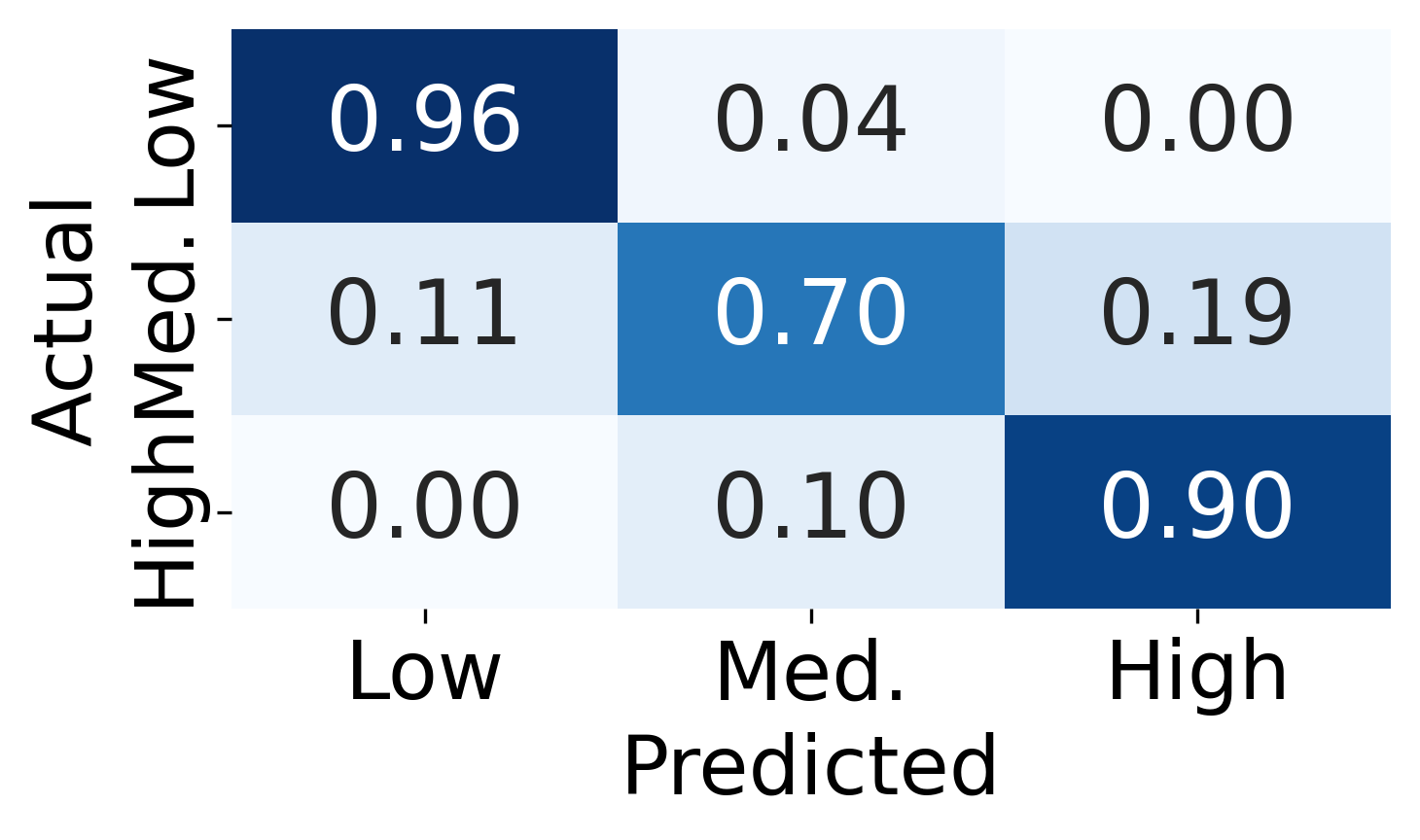}
        \caption{fine-tuned metric split}
        \label{fig:cf_ft_metric}
    \end{subfigure}

    % Second row of subfigures
    \begin{subfigure}[b]{0.48\linewidth}
        \includegraphics[width=\linewidth]{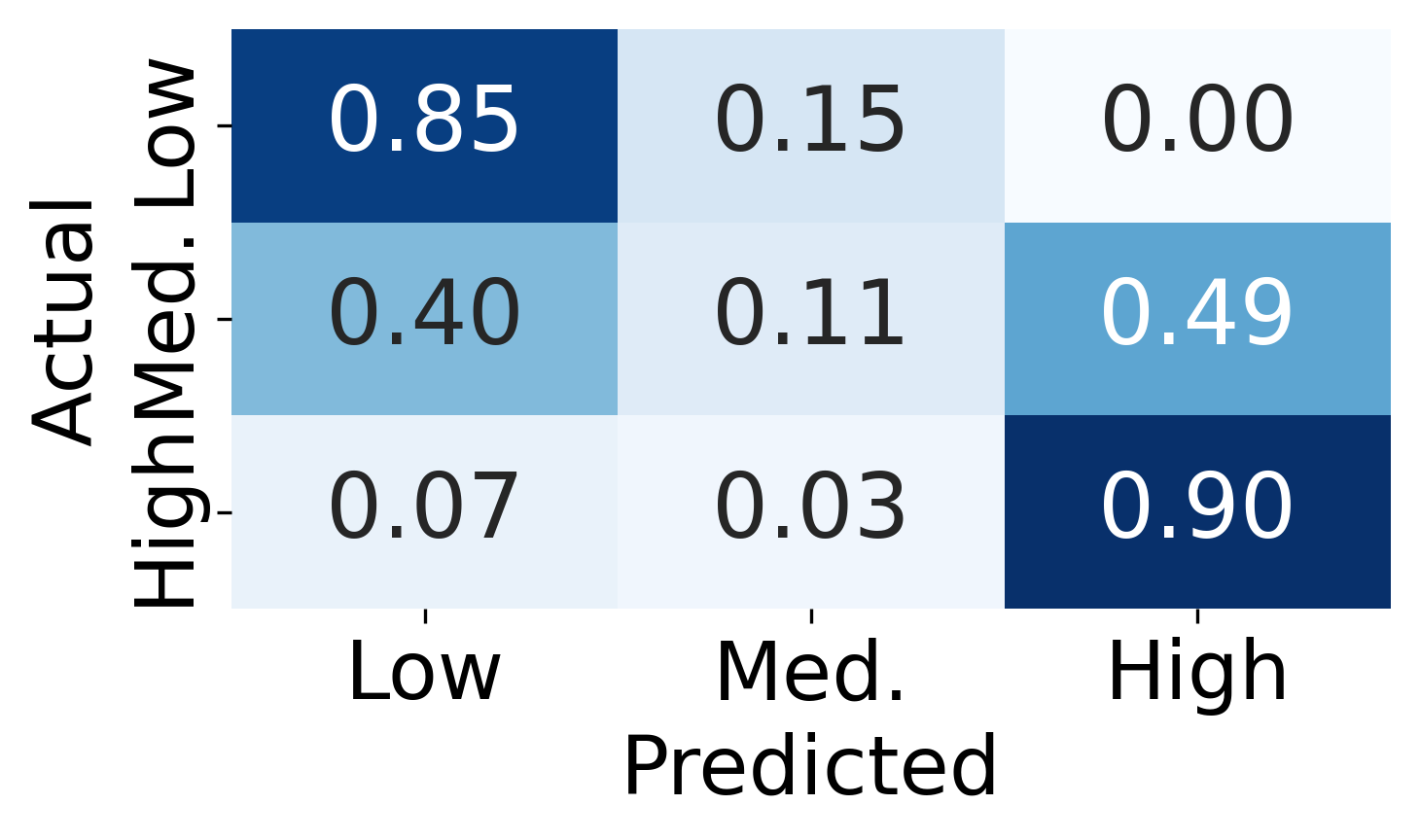}
        \caption{zero-shot question split}
        \label{fig:cf_zh_qs}
    \end{subfigure}
    \hfill % Spacing between the second row subfigures
    \begin{subfigure}[b]{0.48\linewidth}
        \includegraphics[width=\linewidth]{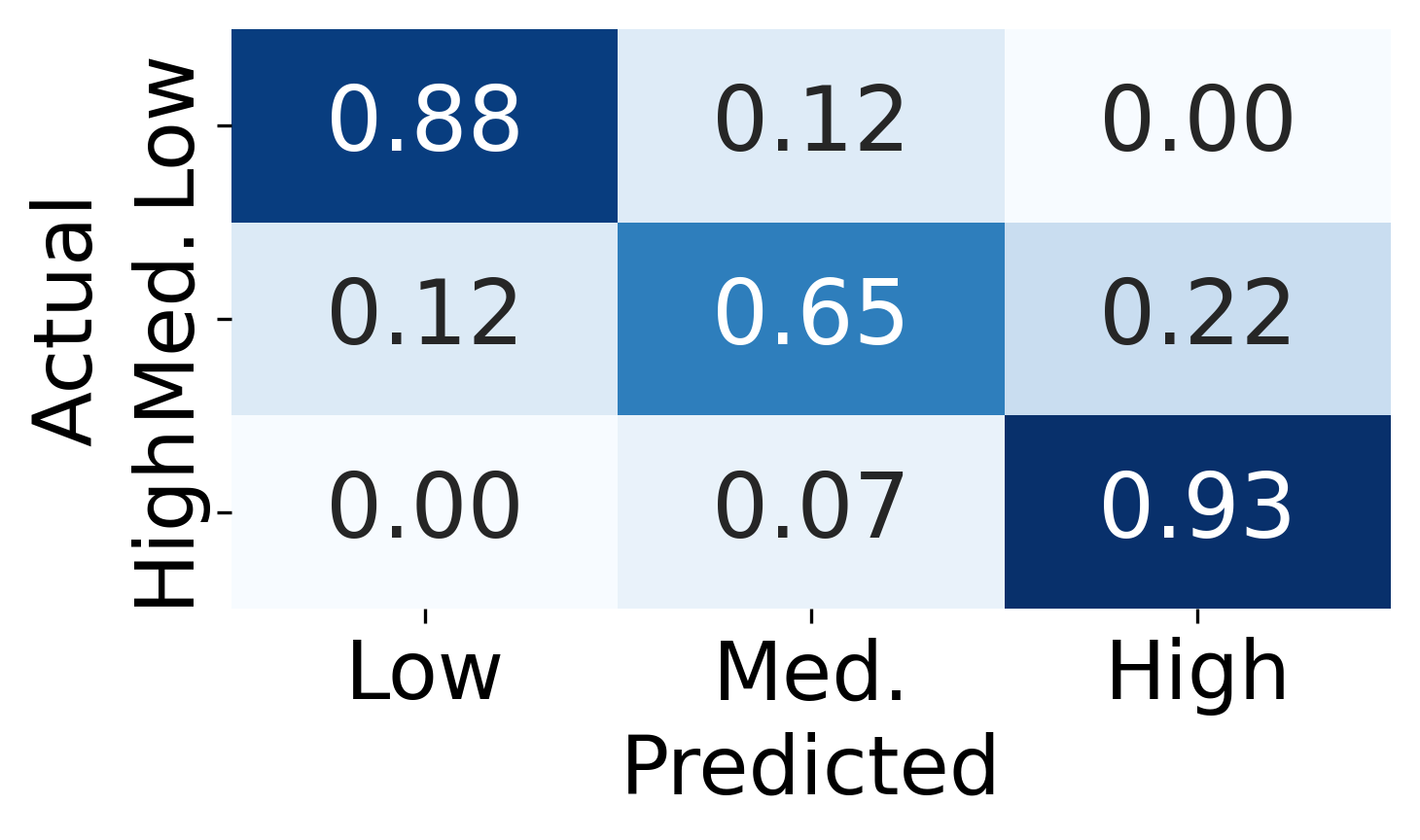}
        \caption{fine-tuned question split}
        \label{fig:cf_ft_qs}
    \end{subfigure}

    % Main caption for the entire figure
    \caption{Confusion matrices for Llama 3 70B Instruct for metric split: zero-shot model (a), fine-tuned (b); and question split: zero-shot (c) and fine-tuned (d).}
    \label{fig:confusion_matrices_qs}
\end{figure}

Results in Table~\ref{tab:combined_results} show that LLMs possess some ability to evaluate counterfactual explanations even with zero-shot learning, with the GPT-4 reaching 63\% accuracy on metric split and Llama 3 70B Instruct reaching 59\% on question split. All models surpassed the 33\% accuracy expected from random guessing in a three-class task. Fine-tuning significantly improved scores, with the Llama 3 70B Instruct reaching 85\% accuracy on the metric split and the recent but significantly smaller Llama 3.1 8B Instruct matching the result. For question split, the highest-performing model was Llama 3 70B Instruct, which after fine-tuning achieved 81\% accuracy for three class prediction across 8 metrics.

\begin{table}[b]
\centering
% Define the width of each column manually
\begin{tabular}{|>{\centering\arraybackslash}p{2.5cm}|>{\centering\arraybackslash}p{0.9cm}|>{\centering\arraybackslash}p{0.9cm}|>{\centering\arraybackslash}p{0.9cm}|>{\centering\arraybackslash}p{0.9cm}|}
\hline
\multirow{3}{*}{\textbf{Metric}} & \multicolumn{2}{c|}{\textbf{Metric Split}} & \multicolumn{2}{c|}{\textbf{Question Split}} \\ \cline{2-5} 
 & Zero- & Fine- & Zero- & Fine- \\
 & shot & tuned & shot & tuned \\
\hline
Satisfaction & 0.67 & \textbf{0.96} & 0.50 & \textbf{0.88} \\ 
\hline
Consistency & 0.58 & 0.83 & 0.83 & 0.88 \\ 
\hline
Feasibility & 0.79 & 0.96 & 0.54 & 0.67 \\ 
\hline
Understand. & 0.54 & \textbf{1.0} & 0.92 & 1.0 \\ 
\hline
Fairness & 0.50 & \textbf{0.83} & 0.67 & \textbf{1.0} \\ 
\hline
Trust & 0.50 & 0.67 & 0.50 & 0.50 \\ 
\hline
Complexity & 0.42 & \textbf{0.75} & 0.42 & \textbf{0.83} \\ 
\hline
Completeness & 0.33 & \textbf{0.83} & 0.33 & \textbf{0.75}\\ 
\hline
\end{tabular}
\caption{Evaluation of various metrics for Llama 3 70B Instruct model. The largest improvements are highlighted in bold. Each of the accuracy scores is the average score over 4 runs. (Understand. is an abbreviation for Understandability).}
\label{tab:metric_results}
\end{table}
 
The confusion matrices show that after fine-tuning, the best-performing models for both splits (Figure~\ref{fig:confusion_matrices_qs}b, Figure~\ref{fig:confusion_matrices_qs}d) made no errors misclassifying 'low' as 'high' or vice versa, suggesting a high-level understanding of the metrics. Table~\ref{tab:metric_results} highlights accuracy improvements after fine-tuning, with notable gains in Completeness (improving from 33\% to 83\% and 75\% for the metric and question split, respectively), Complexity (from 42\% to 75\% and 83\%), and Understandability, which achieved perfect accuracy. Importantly, Satisfaction showed substantial improvements reaching 96\% for metric split and 88\% for question split. Feasibility and Trust remain challenging for prediction, largely because assessing the feasibility and outcomes of categorical changes is complex and often unclear as to whether it would bring the desired outcome.

\subsection{Modelling Individual Preferences}
Different people's preferences for explanations exhibit significant variability. To explore the effects of this, we conducted an experiment using a dataset based on specific participants' answers rather than sample averages. To ensure that these participants represent different subgroups of participants, t-SNE was used to reduce the dimensionality of the data, and DBSCAN clustering identified the largest clusters.  random participant was selected from each of the four largest clusters. The results of clustering and participant selection can be viewed in Appendix D, Figure D.1.

The selected participants, each from different European countries with educational levels from high school to Master’s degree, ensured a diverse range of viewpoints. One participant’s experience in machine learning further enriched the variety of responses, detailed in Appendix D, Table D.1.

\begin{table} [t]
    \centering
    \begin{tabular}{|c|c|c|} \hline 
         \textbf{Participant}&  \textbf{Zero-shot}&  \textbf{Fine-tuned} \\ \hline 
         A&  0.67&  0.87\\ \hline 
         B&  0.58&  0.66\\ \hline 
         C&  0.69&  0.90\\ \hline 
         D&  0.69&  0.90\\ \hline
    \end{tabular}
\caption{Evaluation accuracy over all metrics for four participants that were selected to represent different subgroups of participants.}
\label{tab:ex5}
\end{table}

For each of these participants, zero-shot evaluation and fine-tuning was carried out using the same procedure as in the previous experiments, but using only the model Llama 3 70B Instruct, as it proved to be the most capable (for hyperparameters see Appendix C, Table C.2). The testing set contained the same question-metric pairs as in the first experiment, but with answers from the specific participant. 

The results of this process varied, with accuracies ranging from 58\% to 69\% for zero-shot evaluation. Table~\ref{tab:ex5} shows that the LLM's predictions improved significantly after fine-tuning, reaching accuracies over all metrics of $\sim\!\!90\%$ for 3 participants. One participant appeared to be less consistent, as the model managed to simulate their answers with an accuracy of only 66\%. This leads to two conclusions: while LLM's biases and preferences can be tuned to match specific participants to a great extent, some participants' preferences prove significantly more difficult to mimic. However, since this comparison only contained 4 participants and 30 explanations, these conclusions should be considered tentative.
\section{Discussion}

The traditional assessment of counterfactual explanations often overlooks human aspects, relying either on inconsistent quantitative metrics (frequently used both within objective function optimization and for evaluation \cite{cheng_dece_2021}) or on user studies that focus on a specific subset of individuals, lacking comparability over time and methods. To address this, we developed a novel dataset of counterfactual explanations, evaluated by human participants, which demonstrated a diverse spread of evaluations across all metrics, highlighting its applicability in different contexts. Utilizing this dataset to fine-tune LLMs demonstrated promising results, achieving an 85\% accuracy, suggesting they can be used to approximate human judgment across various metrics. Furthermore, the zero-shot LLM performance was already notable, achieving up to 63\% accuracy. Our experiments also indicate the potential to fine-tune models to individual experts, to target specific expertise or individual preferences.

%The integration of human cognitive biases into the generation and evaluation of CEs emerged as a significant factor in enhancing their relevance and acceptance. %Metrics such as feasibility, which assesses the practicality of the suggested changes, and consistency, which ensures that all parts of an explanation are coherent, emerged as crucial for predicting human satisfaction. These metrics align with psychological principles that emphasize the importance of understandable and actionable information in decision-making processes. Although, due to the inter-correlation of human responses it might be that not every metric is captured their definition in the desired way.

However, employing LLMs for evaluating counterfactual explanations introduces ethical considerations. There is a risk of reinforcing or introducing biases if the models are not continuously monitored and updated with diverse training data. Furthermore, optimizing explanations to align with model preferences might lead to "gaming" the system, skewing results towards what the model favours rather than enhancing the relevance of the explanations to human users.

A considerable limitation of our study is the dataset size, consisting of only 30 unique counterfactual explanations. A larger dataset would likely enhance model training capabilities. Future work should aim to generate larger datasets using recent counterfactual algorithms \cite{rasouli2024care,codice_domnich_2024,dandl_2024_contrafactuals}. These should be presented in smaller subsets to participants for evaluation, given that a single participant can only assess a limited number of explanations thoroughly.

In the future, the main implication of this work is that a fine-tuned LLM should be applied to evaluate various counterfactual algorithms. Additionally, the model can be iteratively retrained with newer and larger architectures and datasets. With the continuously improving size and capabilities of LLMs, this is likely to lead to further improvements in mimicking human evaluation patterns.

Despite the potential, it is crucial to acknowledge that LLMs do not replace the nuanced insights provided by human evaluations. Instead, they can serve as a complementary tool, enhancing scalability and reducing the resources required for broad assessments across multiple frameworks.
Moreover, we propose exploring the idea of integrating this model within a human-in-the-loop approach to produce a hybrid model that could refine the quality of counterfactual explanations during the generation process, effectively creating an LLM-in-the-loop approach instead of a human \cite{carlo_2024_human_in_loop}, combining the strengths of automated and human evaluations.

\section{Conclusion}
This study aims to advance towards more standardized and human-centric evaluations of counterfactual explanations in AI systems. The development and application of our novel dataset, which captures a broad spectrum of human evaluations, reveals the significant potential of LLMs to mirror human judgment with a high degree of accuracy.

\section{Ethical Statement}
All data were collected without any personal identifiers. The study was approved by The University of Tartu Research Ethics Committee, and participants provided informed consent for their anonymized data to be used for educational and research purposes.

\section{Acknowledgments}
This research was supported by Estonian Research Council Grants PRG1604, the European Union’s Horizon 2020 Research and Innovation Programme under Grant Agreement No. 952060 (Trust AI), the Estonian Centre of Excellence in Artificial Intelligence (EXAI), the Estonian Ministry of Education and Research. 

\bibliography{aaai25}
\appendix
\onecolumn
\begingroup
\include{supplementary}

\endgroup
\end{document}

%% file: supplementary.tex
% Nice Todo box
\setlength{\marginparwidth}{2cm}
\newcommand{\TODO}{\todo[inline]}
%%%%%

\begin{appendices}
\section{Appendix A}
\setcounter{table}{0}
\setcounter{figure}{0}
\renewcommand{\thetable}{A.\arabic{table}}
\renewcommand{\thefigure}{A.\arabic{figure}}

\begin{table}[H]
    \centering
    \label{tab:ap1_2}
    \begin{tabular}{|p{\linewidth}|} \hline 
    Imagine you are in this scenario:
    
    \textbf{"You are a 31-year-old person, who has had one pregnancy, and who has the following medical readings: glucose level: 173 mg/dL, diastolic blood pressure: 82 mm Hg, skin thickness: 48 mm, Insulin: 160 $\mu$IU/ml, body mass index (BMI): 32.8."}
    \hfill \break
    Current outcome: You have an \textbf{increased risk of diabetes.}
    \hfill \break
    \scriptsize Useful context: Normal glucose levels are 70-130, 140-200 is prediabetes, 200+ is diabetes.
    Healthy BMI is 18.5 - 25, 25-30 is considered overweight, 30+ is considered obese.
    Normal diastolic blood pressure is roughly below 80 mm Hg.
    Normal insulin is between 16 and 166 $\mu$IU/ml.
    \hfill \break
    
   \normalsize "To \textbf{not be at increased risk of diabetes} you would need to make the following changes:
    \begin{itemize}
        \item Decrease your \textbf{glucose level} from \textbf{173} to \textbf{130}.
        \item And, increase your \textbf{insulin level} from \textbf{160} to \textbf{181}."
    \end{itemize}
    \\ \hline
    On a scale from 1 (very unsatisfied) to 6 (very satisfied), how \textbf{satisfied} would you be with such an explanation:
    
    \hfill \break
    \normalsize On a scale from 1 (very infeasible) to 6 (very easy to do), how \textbf{feasible} is this explanation:
    
    \scriptsize Feasibility - the actions suggested by the explanation are practical, realistic to implement and actionable. (click to see examples)
    \hfill \break
    
    \normalsize On a scale from 1 (very inconsistent) to 6 (very consistent), how \textbf{consistent} is this explanation:

    \scriptsize Consistency - all parts of the explanation are logically coherent and do not contradict each other. (click to see examples)
    \hfill \break
    
    \normalsize On a scale from 1 (very incomplete) to 6 (very complete), how \textbf{complete} is this explanation:

    \scriptsize Completeness - the explanation is sufficient in explaining how to achieve the desired outcome. (click to see examples)
    \hfill \break
    
    \normalsize On a scale from 1 (not at all) to 6 (very much), how much do you \textbf{trust} this explanation:

    \scriptsize Trust - I believe that the suggested changes would bring about the desired outcome. (click to see examples)
    \hfill \break
    
    \normalsize On a scale from 1 (incomprehensible) to 6 (very understandable), how \textbf{understandable} is this explanation:

    \scriptsize Understandability - I feel like I understood the phrasing of the explanation well. (click to see examples) 
    \hfill \break 
    
    \normalsize On a scale from 1 (very biased) to 6 (completely fair), how \textbf{fair} is this explanation:

    \scriptsize Fairness - the explanation is unbiased towards different user groups and does not operate on sensitive features. (click to see examples)
    \hfill \break
    
    \normalsize On a scale from -2 (too simple) to 0 (ideal complexity) to 2 (too complex), how \textbf{complex} is this explanation: 

    \scriptsize Complexity - the explanation has an appropriate level of detail and complexity - not too simple, yet not overly complex. (click to see examples)
    
    \\ \hline
    \end{tabular}
    \caption{Example of a questionnaire question from the final study}
\end{table}

\begin{table}[H]
    \centering
    \begin{tabular}{|>{\centering\arraybackslash}p{0.07\linewidth}|@{ }>{\centering\arraybackslash}p{0.05\linewidth}|@{ }>{\centering\arraybackslash}p{0.07\linewidth}|>{\centering\arraybackslash}p{0.06\linewidth}|@{ }>{\centering\arraybackslash}p{0.09\linewidth}|>{\centering\arraybackslash}p{0.07\linewidth}|>{\centering\arraybackslash}p{0.07\linewidth}|>{\centering\arraybackslash}p{0.07\linewidth}|>{\centering\arraybackslash}p{0.1\linewidth}|} \hline 
         Partici- pant id&Survey time&  Failed attention check&  PCA outlier&  Similar answering pattern&Question 16& Question 21&  Question 32& Avg. understandability below 3\\ \hline 
         12&  &  x&  x&  x&  x&  x&  x& \\ \hline 
         27&  x&  x&  x&  x&  &  x&  & \\ \hline 
         28&  &  &  &  x&  x&  x&  & \\ \hline 
         44&  &  x&  &  x&  x&  x&  & \\ \hline 
         75&  &  &  &  &  x&  &  & x\\ \hline 
         83&  &  x&  &  x&  x&  x&  & x\\ \hline 
         84&  &  &  x&  x&  &  x&  & \\ \hline 
         88&  &  &  &  x&  x&  x&  x& x\\ \hline 
         97&  &  &  &  x&  x&  x&  x& x\\ \hline 
 201& & & x& & x& & x&\\ \hline
    \end{tabular}
    \caption{Dropped participants and corresponding indicators of low-quality. X marks a check that was failed by the specific participant.}
    \label{tab:drop_table}
\end{table}

\begin{table}[H]
\centering
\renewcommand{\arraystretch}{1.2} % Adjust row height for better readability
\begin{tabular}{@{}p{3cm}lrr@{}}
\toprule
\textbf{Demographic}              & \textbf{Description}                                  & \textbf{Number} & \textbf{Percentage (\%)} \\ \midrule
\multirow{5}{3cm}{\textbf{Age Group}} & 18-24 years old                                 & 64              & 31.07                   \\
                                      & 25-34 years old                                 & 95              & 46.12                   \\
                                      & 35-44 years old                                 & 25              & 12.14                   \\
                                      & 45-54 years old                                 & 15              & 7.28                    \\
                                      & 55-64 years old                                 & 7               & 3.40                    \\ \midrule
\multirow{11}{3cm}{\textbf{Nationality}} & South African                                & 32              & 15.53                   \\
                                         & Polish                                       & 30              & 14.56                   \\
                                         & British                                      & 27              & 13.11                   \\
                                         & Mexican                                      & 20              & 9.71                    \\
                                         & Portuguese                                   & 19              & 9.22                    \\
                                         & Chilean                                      & 12              & 5.83                    \\
                                         & Italian                                      & 8               & 3.88                    \\
                                         & Greek                                        & 7               & 3.40                    \\
                                         & Hungarian                                    & 7               & 3.40                    \\
                                         & USA                                          & 6               & 2.91                    \\
                                         & Other                                        & 38              & 18.45                   \\ \midrule
\multirow{5}{3cm}{\textbf{Education Level}} & Primary school or lower                   & 1               & 0.49                    \\
                                            & High school                               & 69              & 33.50                   \\
                                            & Bachelor's degree or equivalent           & 93              & 45.15                   \\
                                            & Master's degree or equivalent             & 39              & 18.93                   \\
                                            & Doctoral degree or equivalent             & 4               & 1.94                    \\ \midrule
\multirow{2}{3cm}{\textbf{English Proficiency}} & Native speaker / Fully proficient     & 166             & 80.58                   \\
                                                & Moderately proficient                  & 40              & 19.42                   \\ \midrule
\multirow{4}{3cm}{\textbf{Experience in Machine Learning}} & No experience               & 100             & 48.54                   \\
                                                           & Some experience             & 89              & 43.20                   \\
                                                           & I am studying in a related field & 12           & 5.83                    \\
                                                           & I work in the field / Extensive experience & 5 & 2.43                    \\ \midrule
\multirow{3}{3cm}{\textbf{Experience with Causality Frameworks}} & No                    & 145             & 70.39                   \\
                                                                 & I am familiar with the general concept & 57 & 27.67                   \\
                                                                 & I have previous experience with them & 4 & 1.94                    \\ \midrule
\multirow{5}{3cm}{\textbf{Medical Background}} & No                                     & 181             & 87.86                   \\
                                               & I am currently studying medicine or a related field & 10 & 4.85                    \\
                                               & I work with medical data or in a field related to medicine & 8 & 3.88                    \\
                                               & I have a degree in medicine or a related field & 4 & 1.94                    \\
                                               & I work in the field of medicine       & 3               & 1.46                    \\ \bottomrule
\end{tabular}
\caption{Demographics of the participants.}
\label{tab:demographics_full}
\end{table}

\begin{figure} [H]
\begin{center}
\includegraphics[width=0.6\columnwidth]{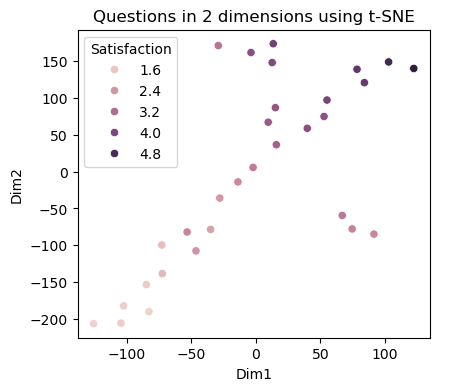}
\caption{Questionnaire questions’ 7 average metric values (no Overall Satisfaction) reduced to 2 dimensions (t-SNE perplexity 3). Colored by average Overall Satisfaction}
\label{fig:t-SNE_questions}
\end{center}
\end{figure}

\newpage
\section{Appendix B}
\renewcommand{\thetable}{B.\arabic{table}}
\renewcommand{\thefigure}{B.\arabic{figure}}
\setcounter{table}{0}
\setcounter{figure}{0}
\label{appendix_examples}
\textbf{Baseline system prompt:}
You are evaluating counterfactual explanations generated by AI. Counterfactual explanations explain what parameters of a situation should have been different for the outcome to have been different. You are not expected to provide reasoning or explanation and should answer with the appropriate value from the set ["low", "medium", "high"]. The definition of completeness: the explanation is sufficient in explaining how to achieve the desired outcome. The following is the counterfactual explanation. 

\textbf{System prompt with all definitions:}
You are evaluating counterfactual explanations generated by AI. Counterfactual explanations explain what parameters of a situation should have been different for the outcome to have been different. You are not expected to provide reasoning or explanation and should answer with the appropriate value from the set ["low", "medium", "high"].
The definition of satisfaction: this scenario effectively explains how to reach a different outcome. 
The definition of feasibility: the actions suggested by the explanation are practical, realistic to implement and actionable. 
The definition of consistency: the parts of the explanation do not contradict each other. 
The definition of completeness: the explanation is sufficient in explaining how to achieve the desired outcome. 
The definition of trust: I believe that the suggested changes would bring about the desired outcome. 
The definition of understandability: I feel like I understood the phrasing of the explanation well. 
The definition of fairness: the explanation is unbiased towards different user groups and does not operate on sensitive features. 
The definition of complexity: the explanation has an appropriate level of detail and complexity - not too simple, yet not overly complex. The following is the counterfactual explanation.  

\textbf{System prompt with examples:}
You are evaluating counterfactual explanations generated by AI. Counterfactual explanations explain what parameters of a situation should have been different for the outcome to have been different. You are not expected to provide reasoning or explanation and should answer with the appropriate value from the set ["low", "medium", "high"]. 
The definition of satisfaction: this scenario effectively explains how to reach a different outcome. 
The definition of feasibility: the actions suggested by the explanation are practical, realistic to implement and actionable. 
The definition of consistency: the parts of the explanation do not contradict each other. 
The definition of completeness: the explanation is sufficient in explaining how to achieve the desired outcome. 
The definition of trust: I believe that the suggested changes would bring about the desired outcome. The definition of understandability: I feel like I understood the phrasing of the explanation well. 
The definition of fairness: the explanation is unbiased towards different user groups and does not operate on sensitive features. 
The definition of complexity: the explanation has an appropriate level of detail and complexity - not too simple, yet not overly complex. 
Here are two examples of a prompt and the output. 
Example prompt 1: "Imagine you are in this scenario: ’You are a 21-year-old person who has an average grade of B. You work part-time for 20 hours per week.’ Current outcome: Your university application was rejected. ’To have your application approved, you would need to make the following changes: Improve your average grade from B to A.’ The rest of the values will remain constant. Please rate as ’low’,’medium’ or ’high’, how consistent is this explanation: " Example output 1: "high". 
Example prompt 2: "Imagine you are in this scenario: ’You are a 21-year-old person who has an average grade of B. You work part-time for 20 hours per week.’ Current outcome: Your university application was rejected. ’To have your application approved, you would need to make the following changes: Increase your hours worked per week from 20 to 80.’ The rest of the values will remain constant. Please rate as ’low’,’medium’ or ’high’, how feasible is this explanation: " Example output 2: "low". Please answer questions in a similar format. The following is the counterfactual explanation.

\begin{table} [ht]
    \centering
\label{tab:2}
    \begin{tabular}{|>{\centering\arraybackslash}p{0.18\linewidth}|>{\centering\arraybackslash}p{0.2\linewidth}|>{\centering\arraybackslash}p{0.2\linewidth}|>{\centering\arraybackslash}p{0.2\linewidth}|} \hline 
         Model&  Base prompt&  With all definitions& With examples\\ \hline 
         Mistral 7B Instruct&  0.40&  \textbf{0.41}& 0.36
\\ \hline 
         Llama 2 7B Chat&  \textbf{0.46}&  0.44& 0.37
\\ \hline 
         Llama 3 8B Instruct&  0.56&  \textbf{0.63}& 0.55
\\ \hline 
         Llama 3 70B Instruct&  0.72&  0.70& \textbf{0.75}
\\ \hline 
         \textbf{Average}&  \textbf{0.54}&  \textbf{0.54}& 0.51
\\ \hline
    \end{tabular}
\caption{Accuracies for different prompt-model combinations. The highest accuracy for each model is highlighted in bold.}
\end{table}
\newpage
\section{Appendix C}
%Hyperparameters for reproducibility
\renewcommand{\thetable}{C.\arabic{table}}
\renewcommand{\thefigure}{C.\arabic{figure}}
\setcounter{table}{0}
\setcounter{figure}{0}

\begin{table} [ht]
    \centering
    \begin{tabular}{|>{\centering\arraybackslash}p{0.1\linewidth}|>{\centering\arraybackslash}p{0.12\linewidth}|>{\centering\arraybackslash}p{0.12\linewidth}|>{\centering\arraybackslash}p{0.12\linewidth}|>{\centering\arraybackslash}p{0.12\linewidth}|>{\centering\arraybackslash}p{0.12\linewidth}|>{\centering\arraybackslash}p{0.12\linewidth}|}\hline
 Split& \multicolumn{3}{|c|}{Metric-wise}& \multicolumn{3}{|c|}{Question-wise}\\\hline
         Model&  Llama 3 70B Instruct&  Llama 3 8B Instruct & Llama 3.1 8B Instruct& Llama 3 70B Instruct&  Llama 3 8B Instruct & Llama 3.1 8B Instruct\\ \hline 
         Batch size&  8&  4 & 4& 8& 4&4\\ \hline 
         Learning rate&  0.0002&   0.00005& 0.00005& 0.0001& 0.0002&0.0001\\ \hline
 Epochs& 5&  5& 6& 5& 4&5\\\hline
 Hardware& 2x NVIDIA Tesla A100 80GB& 1x NVIDIA Tesla A100 80GB &1x NVIDIA Tesla A100 80GB & 2x NVIDIA Tesla A100 80GB &1x NVIDIA Tesla A100 80GB &1x NVIDIA Tesla A100 80GB\\\hline
    \end{tabular}
    \caption{Hyperparameters and hardware used for the fine-tuning of LLMs on averaged human ratings.}
    \label{tab:hyperparameters}
\end{table}

\begin{table} [ht]
    \centering
    \begin{tabular}{|c|>{\centering\arraybackslash}p{0.3\linewidth}|}\hline 
         Model&  Llama 3 70B Instruct\\ \hline 
         Batch size&  8\\ \hline 
         Learning rate&  0.0001\\ \hline
 Epochs& 3\\\hline
 Hardware& 2x NVIDIA Tesla A100 80GB\\\hline
    \end{tabular}
    \caption{Hyperparameters and hardware used for the fine-tuning of LLMs on specific participants answers.}
    \label{tab:hyperparameters_2}
\end{table}

\begin{table}[ht]
    \centering
    \begin{tabular}{|c|c|} \hline 
         r&  32\\ \hline 
         alpha& 
    64\\ \hline 
 Data type&NF4\\ \hline 
 Format&4bit\\ \hline\end{tabular}
    \caption{QLoRA parameters used for fine-tuning LLMs.}
    \label{tab:my_label}
\end{table}

\begin{figure}[ht]
    \centering
    % First row of subfigures
    \begin{subfigure}[b]{0.45\linewidth}
        \includegraphics[width=\linewidth]{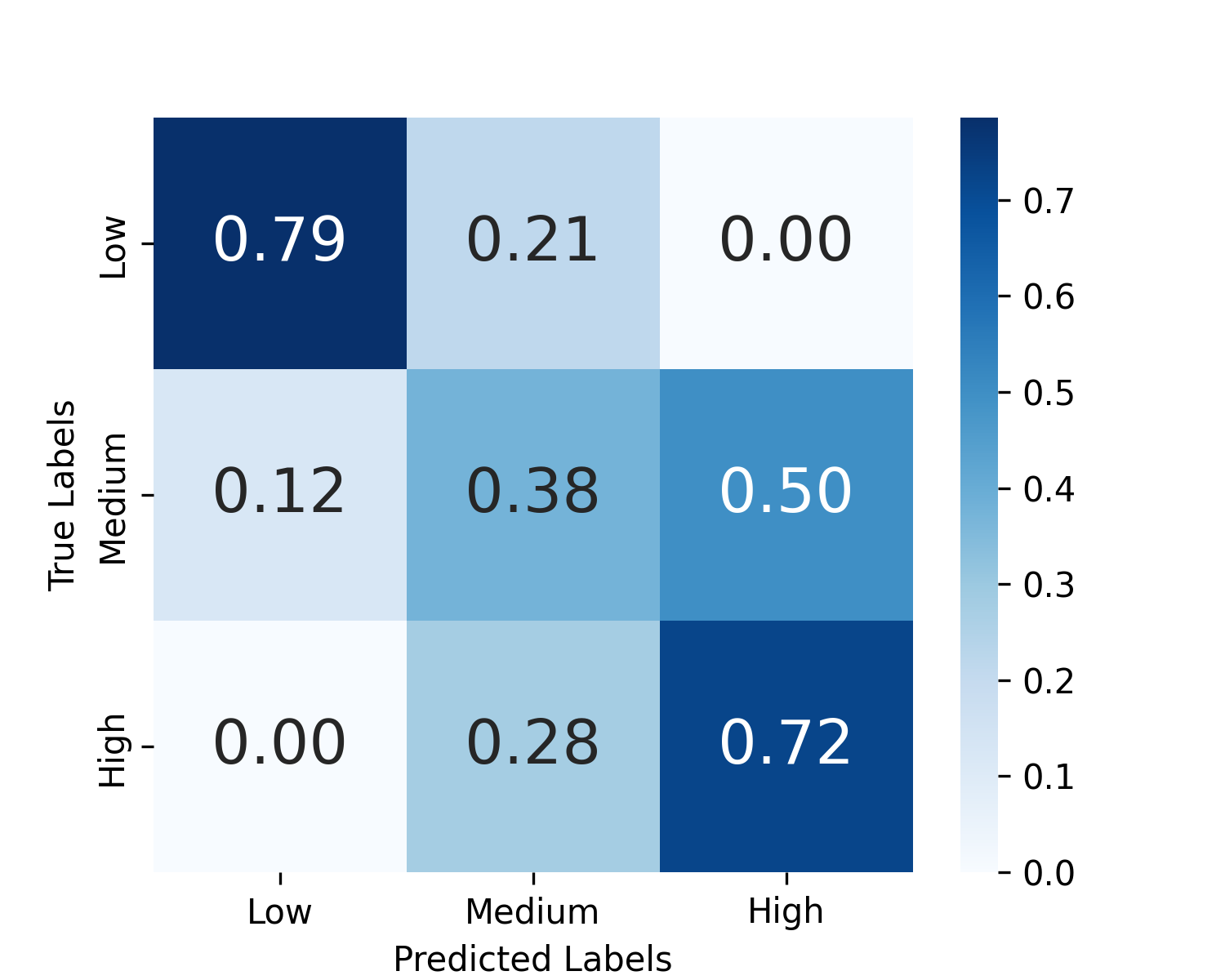}
        \caption{}
        \label{fig:cf_gpt_metric}
    \end{subfigure}
    \hfill % Spacing between the first row subfigures
    \begin{subfigure}[b]{0.45\linewidth}
        \includegraphics[width=\linewidth]{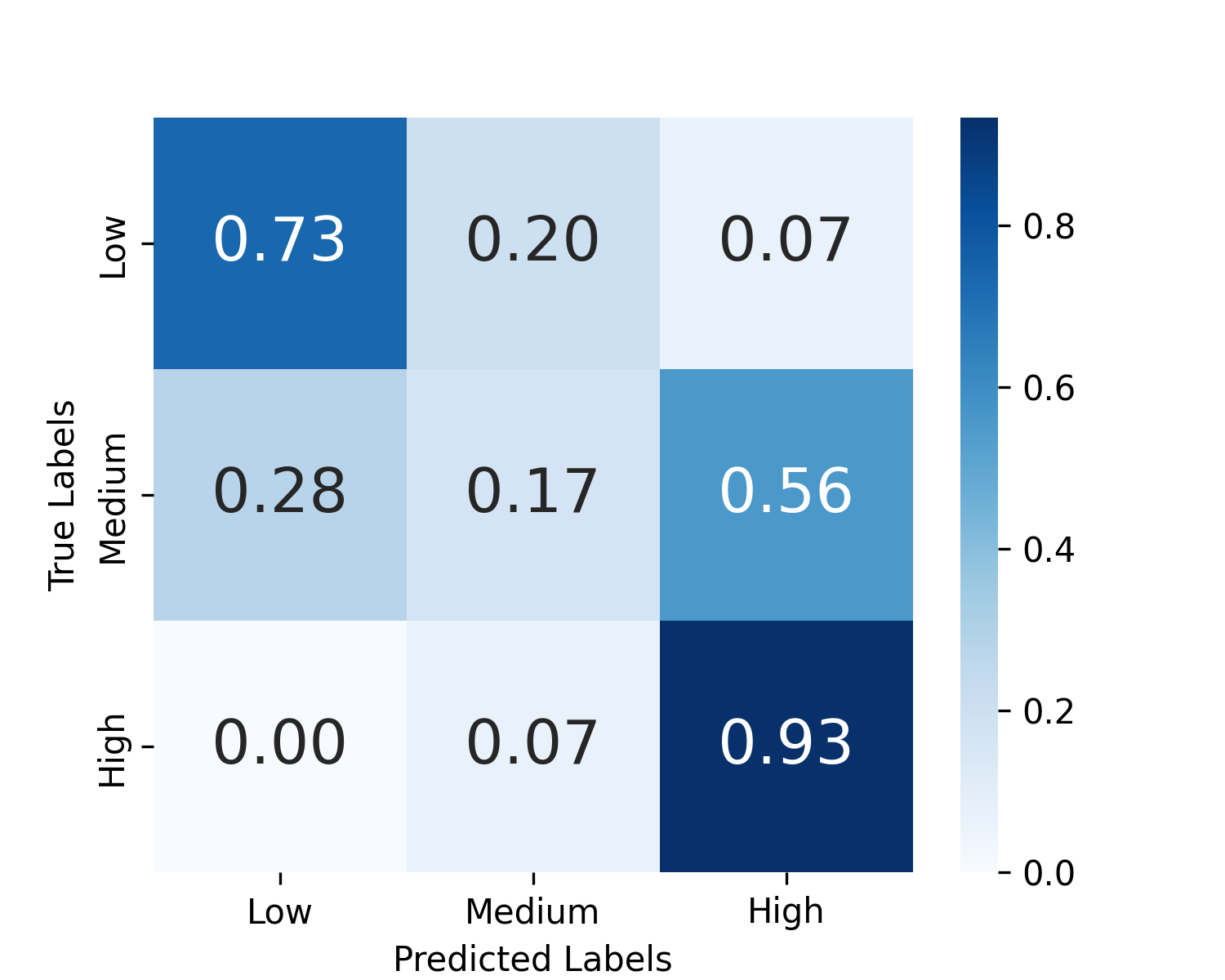}
        \caption{}
        \label{fig:cf_gpt_qsplit}
    \end{subfigure}

    % Main caption for the entire figure
    \caption{Confusion matrices for GPT4 for metric split (a) and question split (b).}
    \label{fig:confusion_matrices_gpt_qs}
\end{figure}

\section{Appendix D}
\renewcommand{\thetable}{D.\arabic{table}}
\renewcommand{\thefigure}{D.\arabic{figure}}
\setcounter{table}{0}
\setcounter{figure}{0}
%This section is about selecting process of individual participants and their background

\begin{figure} [H]
\begin{center}
\includegraphics[width=0.6\columnwidth]{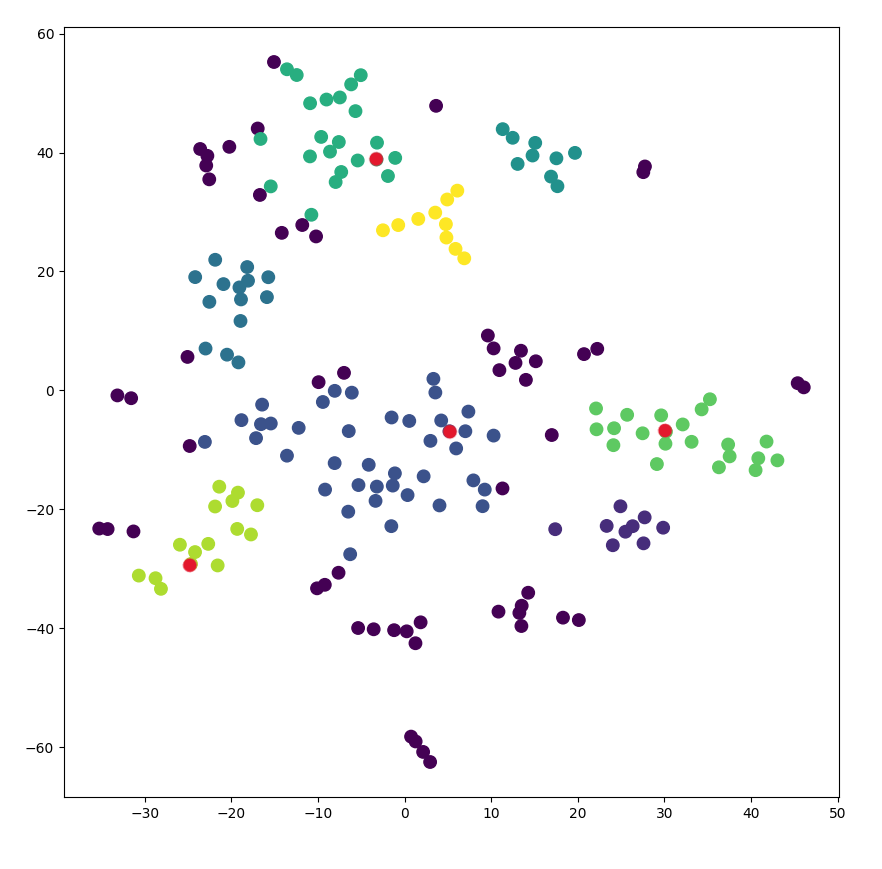}
\caption{DBSCAN clustering of participants. The 4 participants chosen for LLM modelling are marked in red.}
\label{fig:tsne_participants}
\end{center}
\end{figure}

\begin{table*} [ht]
    \centering
    \begin{tabular}{|>{\centering\arraybackslash}p{0.13\linewidth}|>{\centering\arraybackslash}p{0.16\linewidth}|>{\centering\arraybackslash}p{0.16\linewidth}|>{\centering\arraybackslash}p{0.16\linewidth}|>{\centering\arraybackslash}p{0.16\linewidth}|} \hline 
         Participant&  A&  B&  C& D\\ \hline 
         Age&  35-44 years old&  35-44 years old&  25-34 years old&  25-34 years old\\ \hline 
         Citizenship&  Italy&  Portugal&  Poland&  Hungary\\ \hline 
         English proficiency&  Native speaker / Fully proficient&  Native speaker / Fully proficient&  Native speaker / Fully proficient&  Native speaker / Fully proficient\\ \hline 
         Education&  High school&  Bachelor's degree or equivalent&  Master's degree or equivalent&  Master's degree or equivalent\\ \hline
 Experience
with machine
learning& Some experience& No experience& No experience& No experience\\\hline
    \end{tabular}
    \caption{Demographic information of individual participants.}
    \label{tab:demographics}
\end{table*}

\end{appendices}